\documentclass[conference,letterpaper]{IEEEtran}
\IEEEoverridecommandlockouts
\usepackage{graphicx}
\usepackage{multirow}
\usepackage[left=0.71in,top=0.94in,right=0.71in,bottom=1.18in]{geometry}
\usepackage{amsmath}
\usepackage{amssymb}
\usepackage{hhline}
\usepackage{threeparttable}

\begin{document}
\title{\huge Multi-agent Collaboration for Feasible Collaborative Behavior Construction and Evaluation
% $^{*}$\footnoterule
% \thanks{$^{*}$This work is partially
% supported by NSF Grant \#2003168 to H. Simpson and CNSF Grant \#9972988 to M. King}
}

\author{\authorblockN{ First A. Yunkai Wang, Second B. Shenhan Jia, Third C. Zexi Chen, Fourth D. Zheyuan Huang and Fifth E. Rong Xiong}
\authorblockA{\textit{College of Control Science and Engineering}\\
\textit{Zhejiang University}\\
\textit{Hangzhou, Zhejiang Province, China}\\
\textit{\{wangyunkai, 3160104926, chenzexi, 21732054, rxiong\}@zju.edu.cn}\\}%
}

\maketitle

%%%%%%%%%%%%%%%%%%%%%%%%%%%%%%%%%%%%%%%%%%%%%%%%%%%%%%%%%%%%%%%%%%%%%%%%%%%%%%%%
\begin{abstract}
In the case of the two-person zero-sum stochastic game with a central controller, this paper proposes a best collaborative behavior search and selection algorithm based on reinforcement learning, in response to how to choose the best collaborative object and action for the central controller. In view of the existing multi-agent collaboration and confrontation reinforcement learning methods, the methods of traversing all actions in a certain state leads to the problem of long calculation time and unsafe policy exploration. This paper proposes to construct a feasible collaborative behavior set by using action space discretization, establishing models of both sides, model-based prediction and parallel search. Then, we use the deep q-learning method in reinforcement learning to train the scoring function to select the optimal collaboration behavior from the feasible collaborative behavior set. This method enables efficient and accurate calculation in an environment with strong confrontation, high dynamics and a large number of agents, which is verified by the RoboCup Small Size League robots passing collaboration.
\end{abstract}


%%%%%%%%%%%%%%%%%%%%%%%%%%%%%%%%%%%%%%%%%%%%%%%%%%%%%%%%%%%%%%%%%%%%%%%%%%%%%%%%
\section{Introduction}
The research on cooperation and confrontation of multi-agent system (MAS) originated in the 1980s and it is one of the research hotspots in robotics and artificial intelligence. The core problem is to establish a mechanism that enables multiple agents to cooperate with each other to accomplish target tasks and achieve complex intelligence. The research of multi-agent cooperation and confrontation is of great significance, which has been widely used in many fields such as recommendation system, traffic control, unmanned aerial vehicle (UAV) control and game confrontation.\par
Due to the interaction between multiple agents, the complexity of multi-agent system increases rapidly with the increase of number of agents, behavioral complexity and system dynamics. This presents a great challenge to the approach of pre-programming to realize the behavior of multiple agents, which leads to the research of \textit{reinforcement learning (RL)}\cite{RL}, evolutionary computation, game theory, complex system theory and other methods. Among them, reinforcement learning is a hot topic in recent years.\par
In a multi-agent collaboration and confrontation system, there is usually an agent that is most important to achieve the current goal, which is called the central controller or \textit{leader}. A leader usually needs to select an optimal cooperative object and complete the cooperative behavior by executing certain action. For example, in ball games, the leader is a player who holds the ball and needs to select an optimal receiver to complete a passing cooperation by executing the passing action. How to choose the best cooperative object and the appropriate cooperative action is a key problem in multi-agent collaboration and confrontation system. In this paper, the combination of a cooperative object and a cooperative action is defined as \textit{cooperative behavior}, called \textit{COCAP (Cooperative Object-Cooperative Action Pair)}, and an algorithm for searching and selecting optimal cooperative behavior based on RL is proposed. Our method, firstly, according to the action space sampled with a certain precision and models of both sides, predicts the behaviors and final results of all the agents after each cooperative behavior executed, then constructs a set of all feasible collaborative behaviors, finally the best one will be selected by a scoring function which is trained by RL method.\par
In this paper, the \textit{RoboCup Small Size League (SSL)} platform\cite{ssl} is used for verification. It is a centralized and distributed hybrid system, where two teams play against each other and players collaborate to score goals. In essence, it is a two-person zero-sum stochastic game process. The two-person zero-sum stochastic game process refers to that the two parties involved in the game are in a strict competitive relationship, and each time the two parties choose an action, they will gain benefits according to the current state and the selected action. Then a new round of games is carried out in the next random state. The distribution of the new random state depends on the previous state and the actions chosen by both parties. The characteristics of the RoboCup SSL platform are high dynamics and high antagonism, which has certain universality for the research of multi-agent collaboration and confrontation. In this paper, an example of RoboCup SSL robots dynamic passing is presented to solve the problem of multi-robot cooperative passing in high dynamic environment.\par
The remainder of the paper is organized as follow: section \uppercase\expandafter{\romannumeral2} introduces the research of solving the cooperative and confrontative problems of multiple agents by learning method. Section \uppercase\expandafter{\romannumeral3} introduces the best cooperative strategy learning algorithm based on RL. In section \uppercase\expandafter{\romannumeral4}, the proposed algorithm is verified with the RoboCup SSL platform. Section \uppercase\expandafter{\romannumeral5} draws a conclusion, which completes the paper.

\section{Related Work}
In using RL method to solve the problem of multi-intelligence cooperation and confrontation, problems can be classified according to the types of game tasks, namely, fully cooperative, fully competitive and mixed tasks. Fully cooperative task means that all agents can benefit from cooperation without competition. Fully competitive task is a kind of zero-sum game, that is, the gain of one party necessarily means the loss of the other party, and there is no possibility of cooperation. Mixed task means that there is both collaboration and competition between agents. Although they are different problems, the solutions of these three problems have strong similarity and interoperability when using RL method to solve them.\par
In the unique case of optimal joint action, it can be solved by greedy algorithm, that is, each agent chooses the optimal action in the current state, such as \textit{Team Q-learning} method\cite{Team Q-learning} proposed by Littman. When the optimal joint action is not unique, it can be solved by the coordination mechanism between multiple agents, including direct and indirect coordination mechanism. In terms of direct coordination mechanism, Foerster et al.\cite{DIAL} proposed that \textit{Reinforced Inter-Agent Learning (RIAL)} and \textit{Differentiable Inter-Agent Learning (DIAL)} can be used to learn and communicate end-to-end in a complex environment by means of deep q-learning and back-propagation error derivative of communication channel. Sukhbaatar et al.\cite{CommNet} proposed \textit{CommNet}, which enables multiple agents to complete the task of fully cooperation through continuous communication, and enables them to learn communication while learning strategies. Peng et al.\cite{BiCNet} proposed a \textit{Multiagent Bidirectionally-Coordinated Network (BiCNet)} for communication, which builds a actor-critic framework to learn actions and achieve multi-agent cooperation and mastery of various battles in StarCraft games. Indirect coordination mechanism can use the method of agent modeling. For example, Claus et al.\cite{JAL} proposed the \textit{joint action learners (JALs)} which are agents that learn Q-values for joint actions as opposed to individual actions and each agent will model for other agents and combine the model to improve the returns of state action pairs. In the case of confrontation can also use the minimax principle to evaluate the optimization, such as Littman proposed \textit{minmax-Q algorithm}\cite{minmax-Q}, assuming that the opponent will take actions to minimize our benefits and maximize their own.\par
When solving cooperative and confrontative problems of multiple agents by RL, convergence and rationality of learning are very important. Convergence refers to whether the process of agent strategy training is stable and convergent, while rationality refers to the fact that agent's strategy can always converge to an optimal response strategy relative to other players' strategies. In order to achieve convergence and rationality, single-agent algorithm is useful to solve multi-agent problems, such as \textit{WoLF-PHC}\cite{WoLF-PHC} algorithm proposed by Bowling et al. It contains the average strategy of each agent with the current strategy and the current strategy will refer to the average strategy and update the average strategy. The method of centralized learning plus decentralized execution is also used to improve the robustness of the algorithm, such as the \textit{MADDPG}\cite{MADDPG} algorithm proposed by Lowe et al., which enables multiple agents to find complex coordination strategies in physical and information in the cooperative and confrontative environment. And the \textit{Deep-MAHHQN}\cite{Deep-MAHHQN} algorithm proposed by Fu et al. can not only adapt to the discrete-continuous mixed action space, but also show better effect than the independent parameter learning method.\par
Many researchers carry out agent decision-making or multi-intelligent cooperative and confrontative research based on RoboCup SSL platform. In the aspect of agent decision-making, Yoon et al.\cite{Yoon} realized the different shooting skills of robots by means of temporal-difference learning and \textit{multi-layer perceptron (MLP)}. Schwab et al.\cite{Schwab} used \textit{DDPG} algorithm to realize skills learning such as ball finding and shooting, and transferred from simulation to actual robots. In the aspect of cooperation and confrontation between multiple agents, Nakanishi et al.\cite{Nakanishi} studied and analyzed the cooperative passing and shooting task of three robots, and achieved a high success rate in the test of actual robots. Trevizan et al.\cite{Trevizan} used the method of machine learning to propose a similarity function, which compares two teams by imitating the behavior of another team, so as to formulate their strategies. This function can classify opponents and decompose an unknown opponent into a combination of known opponents. Mendoza et al.\cite{Mendoza} proposed a \textit{Selectively Reactive Coordination (SRC)} algorithm to achieve offensive coordination of multiple robots. Behzad et al.\cite{Parsian} used neural network to train opponent model to predict opponent's movement through match recording, so as to get the weakness of opponent's strategy.\par
Currently, to realize multi-agent cooperation and confrontation, RL methods usually traverse all actions in a certain state to obtain the action that maximizes the value function, which is generally approximated by deep neural network with stronger representational ability. However, in most cases, a large number of actions need to be traversed, and the deep neural network takes a long time to operate. The method of traversing all actions will make the calculation time longer, and unsafe actions will be selected in the policy exploration. In this paper, combined with the agent's model, only the behaviors in the feasible cooperative behavior set are traversed and explored, which can not only ensure the feasibility of the selected behaviors in policy exploration, but also reduce the time complexity of the algorithm.\par

\section{Best Collaborative Strategy Learning}
In response to how to choose the best collaborative object and the best collaborative action for the central controller in the case of the two-person zero-sum stochastic game, this paper proposes a best collaborative behavior search and selection method based on RL. Firstly, we discretize the action space, then establish models of both sides and parallel search a feasible collaborative behavior set including actions and collaborative objects. Finally, we degenerate the problem into a Markov decision process, and use the deep q-learning method in RL to train the scoring function for selecting the optimal collaboration behavior from the feasible collaborative behavior set.
\subsection{Feasible Collaborative Behavior Set Search}
For searching a feasible collaborative behavior set, we predict the behavior of the remaining agents and final outcome after the leader performs each collaborative behavior. We use min-max principle\cite{minmax-Q} to predict the result, i.e., assuming that all agents act according to the optimal strategy model after the leader performs a collaborative behavior. Using the optimal strategy model to directly solve or deduct, we can get the prediction results in the future. And actually, in many multi-agent collaboration and confrontation systems, both sides of the confrontation typically use a substantially similar optimal strategy model.\par
Based on the optimal strategy model, we samples the collaborative action space of the leader according to a certain precision, and obtains the collected collaborative action set $A=\{a_i\}$, where $a_i$ represents different collaborative actions. Assuming that all the agents use the optimal strategy to cooperate or confront, and rely on the centralized parallel computing to predict the final result and actions executed by each agent after the leader executes each cooperative action $a_i$. Then we select all of the feasible (or beneficial to us) collaborative behaviors. In this paper, we defined a collaborative behavior consists of collaborative actions and a collaborative object as \textit{``Collaborative Actions-Collaborative Object Pair''(CACOP)}, and collaborative behaviors forms a feasible collaborative behavior set $B=\left\{ \left \langle a_j,r_p \right \rangle \right\}$, where $r_p$ represents different collaboration objects. After that, we calculate the feature vector $x_k$ of each CACOP which may be selected by the leader from the feasible collaborative behavior set. The feature vector $x_k$ is extracted manually that is beneficial to the goal of MAS. Finally, the scoring function was used to calculate scores with respect to the feature vectors. The CACOP with the highest score was selected, and its action was selected as the best cooperative action, and its cooperative object was selected as the best cooperative object.\par
For the passing problem, the optimal strategy of a single robot can be defined as intercepting the moving ball with the maximum movement ability and the shortest time, i.e.
\begin{equation}
\begin{aligned}
& \min~t \\
&{\rm{~s.t.~}} v \leqslant v_{max} \\
&~~~~~~~~ a \leqslant a_{max}
\end{aligned}
\end{equation}
where $t$ is the time when the robot intercepts the ball, $v$ and $a$ are respectively the velocity and acceleration of the robot at any time, and $v_{max}$ and $a_{max}$ are respectively the maximum velocity and acceleration of the robot. We adopt the bang-bang control method proposed by kalmar-nagy et al.\cite{motion} in 2002 for motion planning. That is, at any moment, the robot will accelerate or decelerate at its maximum acceleration, or it will move at its maximum speed.\par
According to this optimal strategy, the interception time of robot can be predicted. In this paper, the time interval sampling search method is adopted, and the fixed minimum time interval is $\Delta t$ (for example, $1/60$ seconds). After the position and velocity of the robots and the ball are obtained through observation at a certain moment, we can calculate the position $P_i$ that the ball can reach after any time $i\Delta t(i=0,1,2,3,...)$ when it does uniform deceleration linear motion under the action of field friction. Then, we start from $i=0$ to search all points, and predict the time $T_i$ required for some robot to reach the $P_i$ point.
If the condition $T_k \leqslant k\Delta t$ is satisfied after the $kth$ time interval, it is considered that $P_k$ is the best interception point $P_{best}$ of the robot, and $k\Delta t$ is the shortest interception time $T_{best}$ of the robot.
In the practical application of RoboCup SSL, in addition to the prediction of the ball rolling on the field, it is also necessary to predict the position of flip shot ball. That is, before preparing to flip shot the ball, we predict the first and second drop locations and arrival times of the ball after flip shot at a certain speed, as well as the subsequent approximate rolling speed. Therefore, when calculating the interception prediction of the flip shot ball, we just need to add the time of the first two jumps in addition.\par
The cooperative actions of the leader robot include kicking mode $c$, kicking direction $\theta$ and kicking speed $v$. In this paper, $128$ bisect samples were taken from the direction of kicking the ball and $16$ bisect samples were taken from the speed of kicking the ball according to the limitation of the kicking ability of the robot, thus forming the cooperative action set $A=\left\{a_i\right\}$, where $a_i=\left \langle c_i,\theta_i,v_i \right \rangle$.
Then, each kick mode, direction and speed were traversed, and up to $24$ robots' interception information were predicted, so that all feasible CACOP could form a feasible cooperative behavior set $B=\left\{ \left \langle a_j,r_p\right \rangle\right\}$, where $r_p$ represents our different cooperative robots. In order to realize real-time high-performance computing, we use $128\times16\times24$ threads on the GPU for parallel computing.\par
On the passing problem, the eigenvector $x_k$ can be composed of the following quantities: the interception time of the robot, the shooting angle of the passing point, the distance from the passing point to the opponent's goal, the angle of shooting after passing, and the interception time of our robot before any other enemy robot. Distance features and angle features are all normalized in order to balance the numerical values of different dimensional features.\par
Through the above work, the feasible cooperative behavior set and the eigenvector representing each feasible cooperative behavior are calculated.

\subsection{Optimal Collaborative Behavior Scoring Function Learning}
Based on the construction of feasible cooperative behavior set, the selection of best cooperative behavior is further considered.\par
A simple and effective method is linear weighted sum, but there are the following problems. First of all, it is difficult to adjust the weight parameters with this method, and it is also difficult to quantify the effects of different parameters. Secondly, the linear weighting of features has its own problems, and the weight of many features should be non-linear. Therefore, the linear weighted sum method has some limitations.\par
A universal and feasible method to solve the above problems is to use MLP to fit the best pass point scoring function. MLP can use its nonlinear activation function to fit the nonlinear function, and use its large number of nodes and depth to fit the higher-order function.\par
The common training method of MLP is supervised learning, that is, providing MLP input vector and target output vector for supervision, and training network weight by comparing the difference between the output and target output. However, using supervised learning is relatively difficult for the following reasons: First, manual annotation is needed to select an best CACOP. This job sometimes is highly professional. Secondly, manually annotated data need to be stored offline in some format for training and use. In general, supervised learning needs a large amount of data as support, which means that a large amount of data needs to be annotated manually. Therefore, using supervised learning to train the MLP as a scoring function is not ideal.\par
Another way to train MLP is reinforcement learning. The method of RL only needs to provide a relatively obvious reward function without manual labeling, and the method of RL can fully consider the cumulative return, that is, learning the causal relationship in Markov decision process. RL can not only learn offline data, including expert presentation data and opponent generated data, but also learn online to interact with the environment to generate new data. Therefore, this paper uses RL method to train the MLP to obtain the scoring function of cooperative behavior.\par
To use RL for training, first define a Markov decision process. In this paper, each collaboration process is regarded as a state $s$ of Markov decision process, and the quantity describing this state is the eigenvector $x_k$ mentioned above. The action $a$ in Markov decision process is to select different CACOP. In this paper, it is considered that the jump from one collaboration state to another is only related to the previous collaboration state and the selected CACOP, and not related to the previous one. Such a process conforms to the conditions of Markov decision process. For example, for the robot passing problem, in a certain cooperation process, the robot 1 passes the ball to the robot 2, and the state of the cooperation process describing the state of $s$ is: the interception time of the robot 2, the shooting angle of the passing point, the distance from the passing point to the opponent's goal, the angle at which the shooting ball is refracted after passing the ball, and the time when the robot 2 preferentially intercepts the ball than the enemy robot. After completing this collaborative process, we make decisions with robot 2 as leader, select the best cooperative action of the robot 2 and the best collaborative object, and then performs the next collaborative process to obtain the next cooperative state $s^\prime$.\par
On the basis of defining the Markov decision process, it is assumed that the scoring function is $Q(s, a)$, and this value function is updated according to the formula of temporal difference (TD) learning:
\begin{equation}
Q(s,a) \leftarrow Q(s,a)+\alpha(R_{t+1}+\gamma max_{a^\prime} Q(s^\prime,a^\prime)-Q(s,a))
\end{equation}
Among them, $\alpha$ stands for the learning rate of RL, $\gamma$ is the discount factor, $s$ stands for the collaboration status, $a$ stands for the selected CACOP, $s^\prime$ stands for the next collaboration state, $a^\prime$ represents the next selected CACOP, $R_{t+1}$ represents the next immediate reward, and $max_{a^\prime} Q(s^\prime,a^\prime)$ means to select a collaboration behavior $a^\prime$ in the state $s^\prime$ to make $Q(s^\prime , a^\prime)$max.

\section{Experiment}
\begin{figure}
\centering
\includegraphics[height=1.8in]{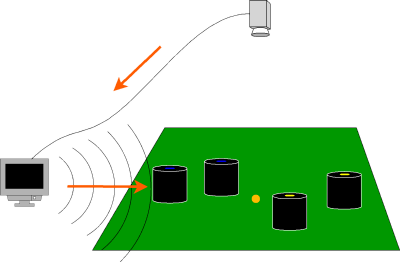}\\
\caption{
Small size robot soccer system architecture
}
\label{ssl}
\end{figure}

This paper uses the RoboCup SSL platform to conduct experiments and verify the effectiveness of the method through ball passing collaboration. Fig.\ref{ssl} is a schematic diagram of the robot system architecture. On the actual playing field, each side has $8$ robots and an orange golf ball as a football. The visual system above the field is processed according to the color of the ball and color code on the top of robots. The number, position and orientation of the robot and the position of the ball are processed by an official dedicated software and sent to each team's host. Then each team's host makes decisions and sends commands to the robots on the field by radio.
The size of the field is $12m$ long and $9m$ wide. The diameter of the robot is $180mm$, and the maximum speed allowed for robot kicking is $6.5m/s$. The motion performance of the robot used in this experiment is shown in the Table \ref{tab1}. In the actual game, the ball-carrying robot can kick the ball at a certain speed in a certain direction by means of a flat shot or a shot, and then the cooperative robot moves to the ball that is calculated in advance to intercept the moving ball, and then passes the ball to the next cooperative robot.\par

\begin{table}
\centering
\caption{Robot performance table}
\label{tab1}
\begin{tabular}{| c | c |}
\hline
\textbf{Performance names} & \textbf{parameters}\\
\hline
Max speed  & $3m/s$ \\
\hline
Max acceleration & $4.5m/s^2$ \\
\hline
Max rotational speed  & $15rad/s$ \\
\hline
Max rotational acceleration & $15rad/s^2$ \\
\hline
\end{tabular}
\end{table}

\begin{figure}[h!t]
\centering
\begin{minipage}{.25\textwidth}
  \centering
  \includegraphics[height=1.3in]{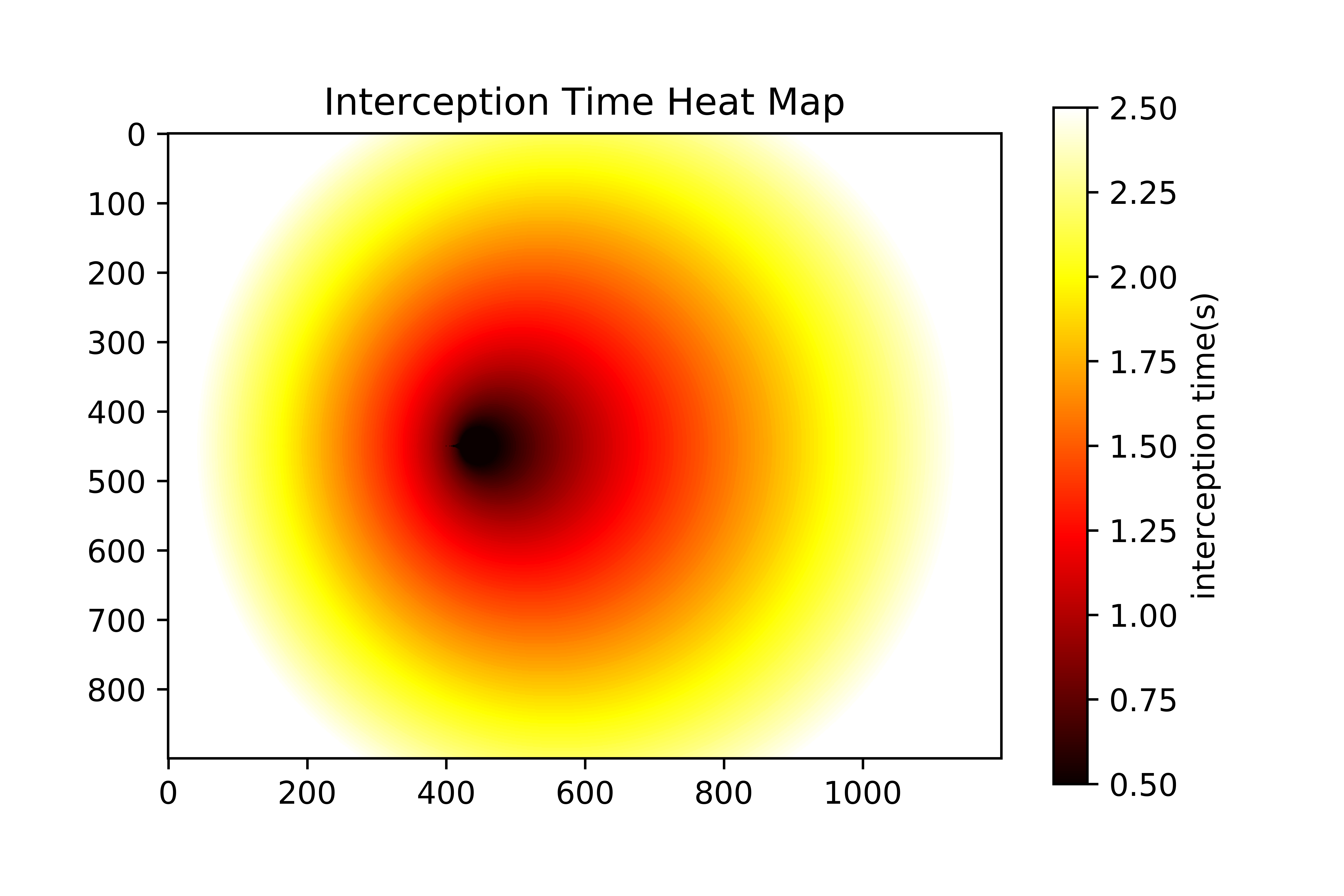}
  \caption{$1m/s$ ball speed interception time heat map. The ball moves from the $(400cm, 450cm)$ to the right at a slower initial speed of $1m/s$.}
  \label{inter_1}
\end{minipage}%

\begin{minipage}{.25\textwidth}
  \centering
  \includegraphics[height=1.3in]{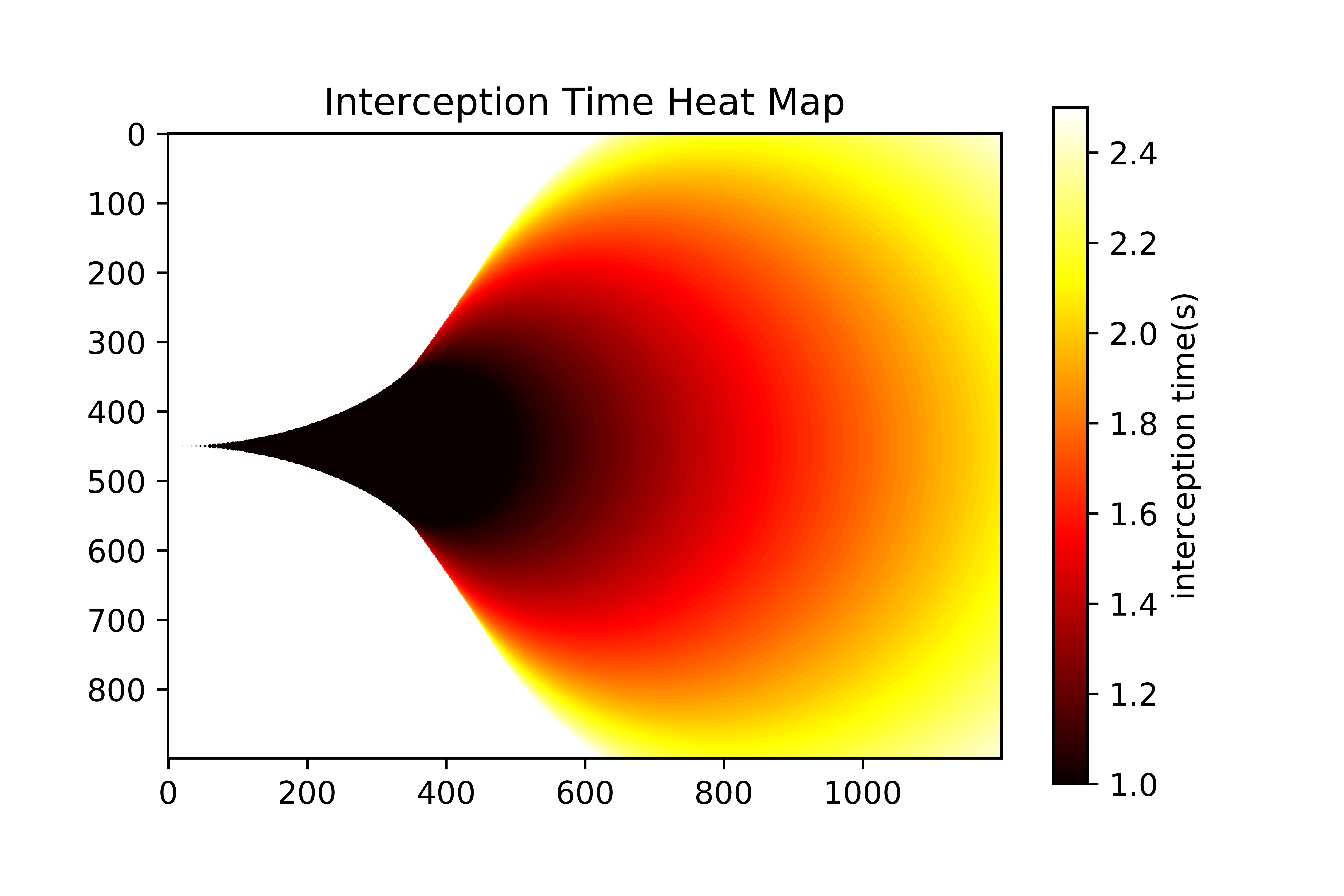}
  \caption{$4m/s$ ball speed interception time heat map. The ball moves from the $(0cm, 450cm)$ to the right at a faster initial speed of $4m/s$.}
  \label{inter_4}
\end{minipage}
\end{figure}

This paper first validates the optimal strategy model for constructing a feasible collaborative behavior set. In the experiment, the interception time of the stationary robot at different positions of the field was tested under the condition of $1m/s$ and $4m/s$. The performance of the robot in the experiment was limited according to the parameters in table \ref{tab1}. The results are shown in Fig.\ref{inter_1} and Fig.\ref{inter_4}. The darker areas of the heat map represent shorter interception times, while the lighter areas represent longer interception times. In Fig.\ref{inter_1}, the ball moves to the right at a slower initial speed of $1m/s$. In this case, the closer the robot is to the ball, the faster it can intercept the ball. However, when the ball speed is faster, there will be different conclusions. As shown in the Fig.\ref{inter_4}, the ball moves to the right at a faster initial speed of $4m/s$, and the robot cannot intercept the ball at the left position. There is a clear boundary in the heat map. If the position of the robot is within this boundary (i.e. dark area), the ball can be intercepted in a short time, while outside the boundary (i.e. white area) it cannot be intercepted in the field.\par

\begin{figure}[h!t]
\centering
\begin{minipage}{.25\textwidth}
  \centering
  \includegraphics[height=1.6in]{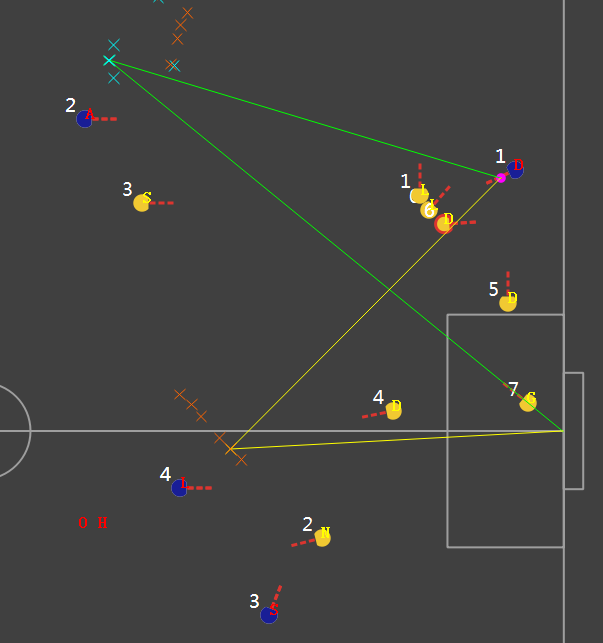}
\end{minipage}
\begin{minipage}{.2\textwidth}
  \centering
  \includegraphics[height=1.6in]{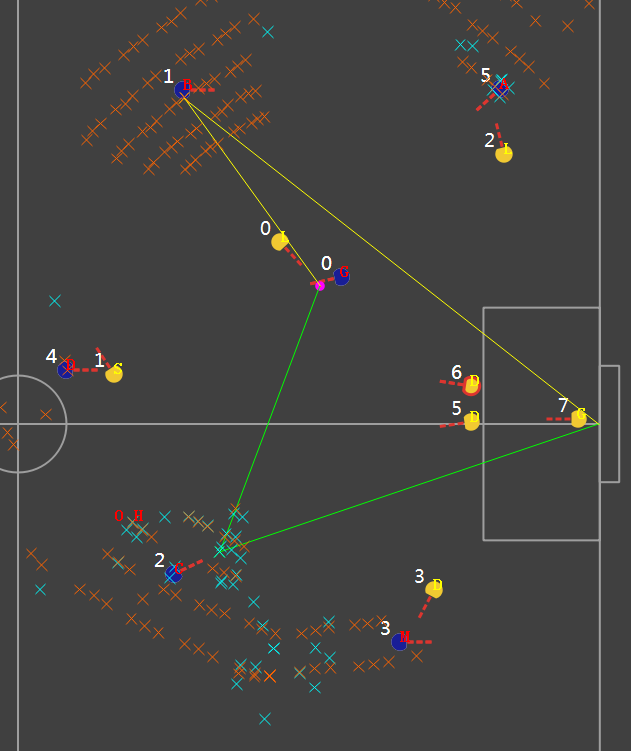}
\end{minipage}
\caption{Visual renderings of feasible collaborative behavior.The light blue points represent feasible flat pass points and the orange points represent feasible flip pass points. The green line represents the best one-shot ball trajectory with a flat shot and the green line represents the best one-shot ball trajectory with a flip shot.}
\label{DPS}
\end{figure}

On the basis of obtaining the optimal strategy model, this paper visualizes the best intercept point for all feasible CACOPs calculated at two different times in the game and the best flat shot and flip shot method obtained by linear weighted sum method, as shown in Fig.\ref{DPS}. The entire process of the algorithm can achieve an average operation speed of $4.04ms$ per frame on the GTX 1060 6G GPU, while using the method for all CACOPs traversal requires an average of $13.77ms$ per frame, which is about $3.4$ for the former. Therefore, the efficiency of the method can be verified. Since the maximum speed of kicking the ball is $6.5m/s$, which is larger than the robot's ability to move, the feasible pass points obtained are mostly in the vicinity of the best cooperative robot. In the case where the enemy robot is still on the field, the pass success rate of almost $100\%$ can be achieved by this method. In the case of enemy robot movement on the field, there may be no feasible flat pass method, but there is always a feasible flip pass method. Using the linear weighted sum method to simply adjust the parameters can give relatively reasonable results, but it is difficult to get a similar effect to manual programming.\par

At the end of the paper, the method of RL is used to train the passing method of scoring function. The reward function mainly consists of two parts: the first part reward $r_1$ is a dense reward based on the distance of the ball from the goal, in the form of
\begin{equation}
r_1 = e^{-\frac{x}{a}}
\end{equation}

Where $x$ represents the distance of the ball from the center of the goal in meters, and $a$ is a constant coefficient. When the ball is very close to the goal, the first part reward is close to $1$, and when the ball is very far from the goal, the first part reward is close to $0$. According to the size of the field, when the ball is $3$ meters away from the center of the goal, the first part reward is $0.5$, so that the solution $a$ is $4.33$. The second part reward $r_2$ is the sparse reward of whether the ball enters their penalty area. When the ball enters their penalty area, $r_2$ is $10$, and when the ball enters the goal, $r_2$ is $50$. Finally, the total reward function is
\begin{equation}
r = r_1 + r_2
\end{equation}
This reward function allows the RL agent to learn how to pass the ball and score a goal.\par
Considering that multiple robots can learn to pass well in the beginning and improve the utilization of data in online RL training, this paper uses all the official game log of the 2018 RoboCup SSL\cite{log} as the empirical data for offline training. After the offline training on the game log, the self-confrontation competition is carried out in the simulation\cite{grSim} to generate more data to train, and then repeat this training process. After a period of training, the trained MLP is obtained, which is the scoring function of the passing.\par
\begin{figure}[hb]
\centering
\includegraphics[width=.45\textwidth]{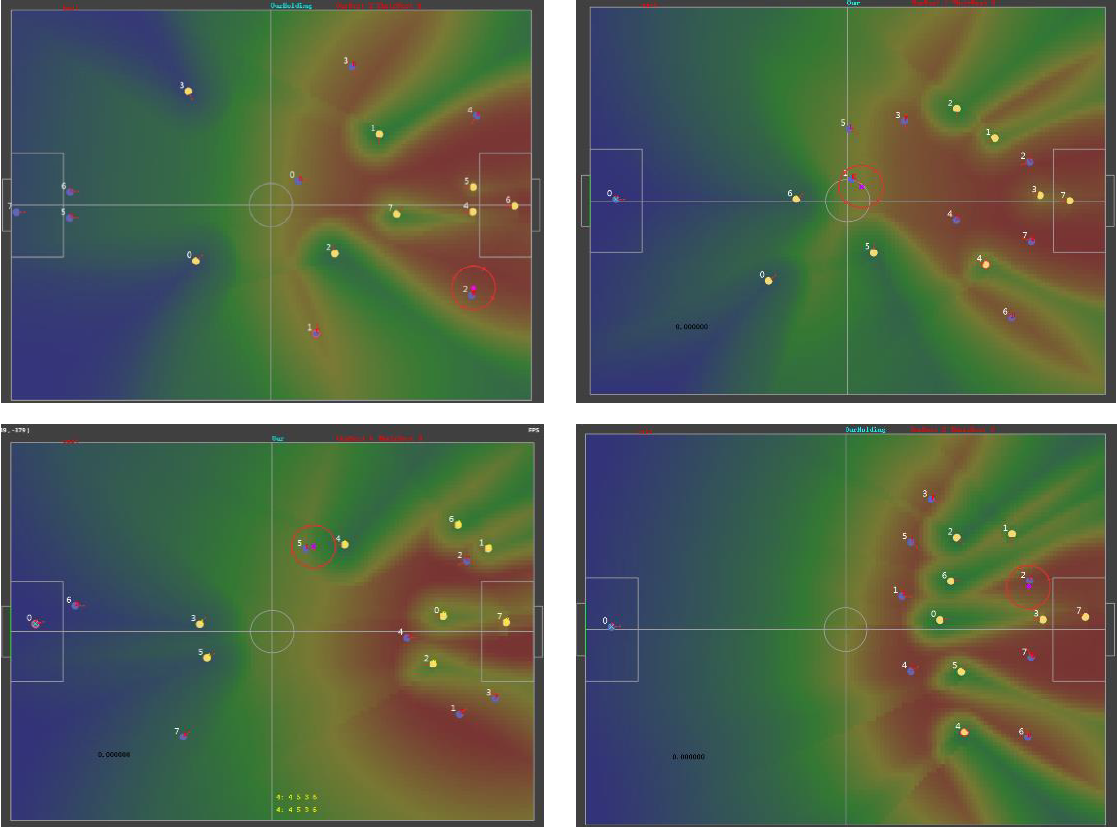}
\caption{The heat map of the pass score obtained by the algorithm. The redder color area represents a higher pass score, and the bluer color area represents a lower pass score. The yellow circle represents the enemy robot, the blue circle represents our robot, and the purple circle represents the ball.}
\label{final}
\end{figure}

Using the trained evaluation function to perform a game in the simulation environment, and visualizing the scores of the best intercept points corresponding to all CACOP obtained by the algorithm, the effect is shown in Fig.\ref{final}.The redder color area represents a higher pass score, and the bluer color area represents a lower pass score. The areas with low scores are mainly concentrated near our half and enemy (yellow) robots, while the areas with high scores are mainly concentrated near the opponent's penalty area and near our (blue) robot, indicating that the selected pass area is in line with human thoughts\par
Finally, this paper carried out 4v4 attack and defense test\footnote{https://youtu.be/S40VmSYvlPks}, that is, 4 robots of the offensive side only pass the ball but not shoot, 4 robots of the defensive side only defend and not grab the ball, to test the rationality of the offensive side pass the ball. The offensive side can pass the ball to the position with high threat degree under high dynamic condition by using the algorithm in this paper, which further verifies the effectiveness of the algorithm.\par

\section{Conclusions}
Aiming at the problem of how to choose the best cooperative object and the best cooperative action in the multi-agent collaboration and confrontation system, this paper proposes a method to search for the best cooperative behavior based on the feasible collaborative behavior set. According to the optimal strategy models of both sides, all feasible cooperative behaviors can be obtained by parallel search on GPU, and the optimal cooperative behaviors can be selected by training score function of RL method. The proposed method is verified by experiments in the RoboCup SSL robots, which is a multi-agent system, and it can make the robots cooperate with other robots in a highly dynamic and confrontational environment, so that the whole system can show high intelligence. It has been applied to actual competitions and relied on the superiority of the algorithm to achieve the 2019 RoboCup SSL world championship.\par
To achieve good cooperation among multiple agents, not only the best cooperation method of the leader should be considered, but also the cooperation of other agents should be considered. For example, in the ball passing problem, when the robot is marked by the opponent, it needs to rely on its own movement to get cooperation opportunities. When the robot marks the opponent's robot in reverse, it needs to prevent the opponent's robot from grabbing the ball through blocking, so that the rest of our robots can pass the ball to the advantageous position of attack. All the receiving robots have no good running position, and the passing robot can not even come up with a great advantage in the way of passing. Therefore, in the future, we can use models of both sides and search methods to get the cooperation method of cooperative agents, so as to achieve higher intelligence in the cooperative behavior of multiple agents.

\end{document}